\documentclass[10pt,twocolumn,letterpaper]{article}
\usepackage{cvpr}              

\usepackage{graphicx}
\usepackage{amsmath}
\usepackage{amssymb}
\usepackage{microtype}
\usepackage{booktabs}
\usepackage{lipsum}
\usepackage{tabularx}
\usepackage{adjustbox}
\usepackage{soul}
\usepackage[accsupp]{axessibility}

\usepackage[pagebackref,breaklinks,colorlinks]{hyperref}

\usepackage[capitalize]{cleveref}
\crefname{section}{Sec.}{Secs.}
\Crefname{section}{Section}{Sections}
\Crefname{table}{Table}{Tables}
\crefname{table}{Tab.}{Tabs.}
\pdfoutput=1

\def\cvprPaperID{7056} 
\def\confName{CVPR}
\def\confYear{2023}
\begin{document}


\newcommand{\mo}[1]{\textcolor{red}{[\texttt{Matt}: #1]}}
\newcommand{\pg}[1]{\textcolor{green}{[\texttt{Pablo}: #1]}}
\newcommand{\gb}[1]{\textcolor{blue}{[\texttt{Gaurav}: #1]}}
\newcommand{\sal}[1]{\textcolor{magenta}{[\texttt{Sally}: #1]}}

\newcommand{\todo}[1]{\textcolor{cyan}{[\texttt{TODO}: #1]}}

\newcommand{\isdraft}{false}


\title{Implicit Neural Head Synthesis via Controllable Local Deformation Fields}

\author{Chuhan Chen$^1$\thanks{Work was done while interning at Flawless AI.}
\and
Matthew O'Toole$^1$
\and
Gaurav Bharaj$^2$
\and
Pablo Garrido$^2$
\and
\begin{tabular}{ c c }
\small{$^1$Carnegie Mellon University} &\small{$^2$Flawless AI}
\end{tabular}
}
\maketitle

\begin{abstract}
High-quality reconstruction of controllable 3D head avatars from 2D videos is highly desirable for virtual human applications in movies, games, and telepresence. 
Neural implicit fields provide a powerful representation to model 3D head avatars with personalized shape, expressions, and facial parts, \eg, hair and mouth interior, that go beyond the linear 3D morphable model (3DMM). 
However, existing methods do not model faces with fine-scale facial features, or local control of facial parts that extrapolate asymmetric expressions from monocular videos. Further, most condition only on 3DMM parameters with poor(er) locality, and resolve local features with a global neural field. 
We build on part-based implicit shape models that decompose a global deformation field into local ones.
Our novel formulation models multiple implicit deformation fields with local semantic rig-like control via 3DMM-based parameters, and representative facial landmarks. 
Further, we propose a local control loss and attention mask mechanism that promote sparsity of each learned deformation field. 
Our formulation renders sharper locally controllable nonlinear deformations than previous implicit monocular approaches, especially mouth interior, asymmetric expressions, and facial details. Project page: \href{https://imaging.cs.cmu.edu/local_deformation_fields/}{ https://imaging.cs.cmu.edu/local\_deformation\_fields/}\looseness=-1
\end{abstract}


\section{Introduction}
\label{sec:introduction}
Monocular human head avatar reconstruction is a long standing challenge that has drawn a lot of attention in the last few decades due to its wide application in movie making \cite{fyffe2014_driving-hd-scans, guo2019_relightables, riviere2020_single-shot}, and virtual reality \cite{bi2021_deep-relightable, ma2021_pixel-codec, cao2022_authentic}, among others. Traditional reconstruction methods in production pipelines create animatable and detailed avatars, often represented as 3D rigs, from high-quality face scans with predefined expressions and poses~\cite{vonderpahlen_digital-ira, wang2020_facial-exp}. However, such data is often expensive to acquire and process, and over the years has created the need for an easier capture pipeline, \eg, based on high-definition images, or videos of human subjects. With the advancements in deep learning, much effort has gone into learning neural 3D face representations from 2D images and the research community has achieved impressive results \cite{lattas2020_avatarme,br2021_learn-complete, lin2020_towards-hf,sharma2022_3d-face-survey}. However, modeling 3D structures from 2D information alone is an ill-posed problem, which results in models that lack view consistency and details.

Both traditional and neural reconstruction pipelines based on the parametric mesh representation, 3DMM~\cite{egger2020_3dmfm}, are efficient, controllable, and well integrated into the graphics pipeline, though at the expense of lacking important facial features such as hair, eyes, and mouth interior. In the last couple of years, there has been a surge of research on generalized implicit face representations, \eg, sign distance functions (SDFs) \cite{or-el2022_stylesdf}, neural radiance fields (NeRFs) \cite{gu2022_stylenerf} or hybrid volumetric representations \cite{chan2022_efficient-geom-aware}, that allow accurate modeling of fine-grained facial, and non-facial features not possible with mesh-based representations alone, while preserving view-consistency.

Recently, several implicit models for human head avatars from monocular videos have demonstrated great progress \cite{zheng2021_imavatar, zheng2022_imface, ren2022_facial, gafni2021_dynamic-nerf, athar2022_rignerf, park2021_nerfies, hong2021_headnerf, wang2022_morf, cao2022_authentic, sun2022_controllable, grassal2021_neural-head, park2021_hypernerf}. Several employ facial parameters from an existing face tracker to condition a multi-layer perceptron (MLP) to model expression changes \cite{gafni2021_dynamic-nerf, hong2021_headnerf, sun2022_controllable, cao2022_authentic}, use deformation fields to learn a mapping from an observed point deformed by expression to a point on the template face \cite{zheng2022_imface, park2021_nerfies, athar2022_rignerf}, or learn forward mapping functions that estimate implicit 3DMM functions, represented as facial deformation bases \cite{zheng2021_imavatar, ren2022_facial-geom, grassal2021_neural-head}. These approaches have done a good job in allowing control of expressions and poses, even for out-of-distribution training parameters. However, none of these approaches reconstruct animatable heads with high-fidelity details such as deep creases. Besides, since they heavily rely on creating implicit fields derived from linear 3DMMs, which are de-facto limited by global or large-scale expression decompositions, it is relatively difficult to control localized deformations at a finer level, \eg, wrinkles that form around eyes when winking.

\begin{figure*}[ht]
    \centering
    \includegraphics[width=0.95\linewidth]{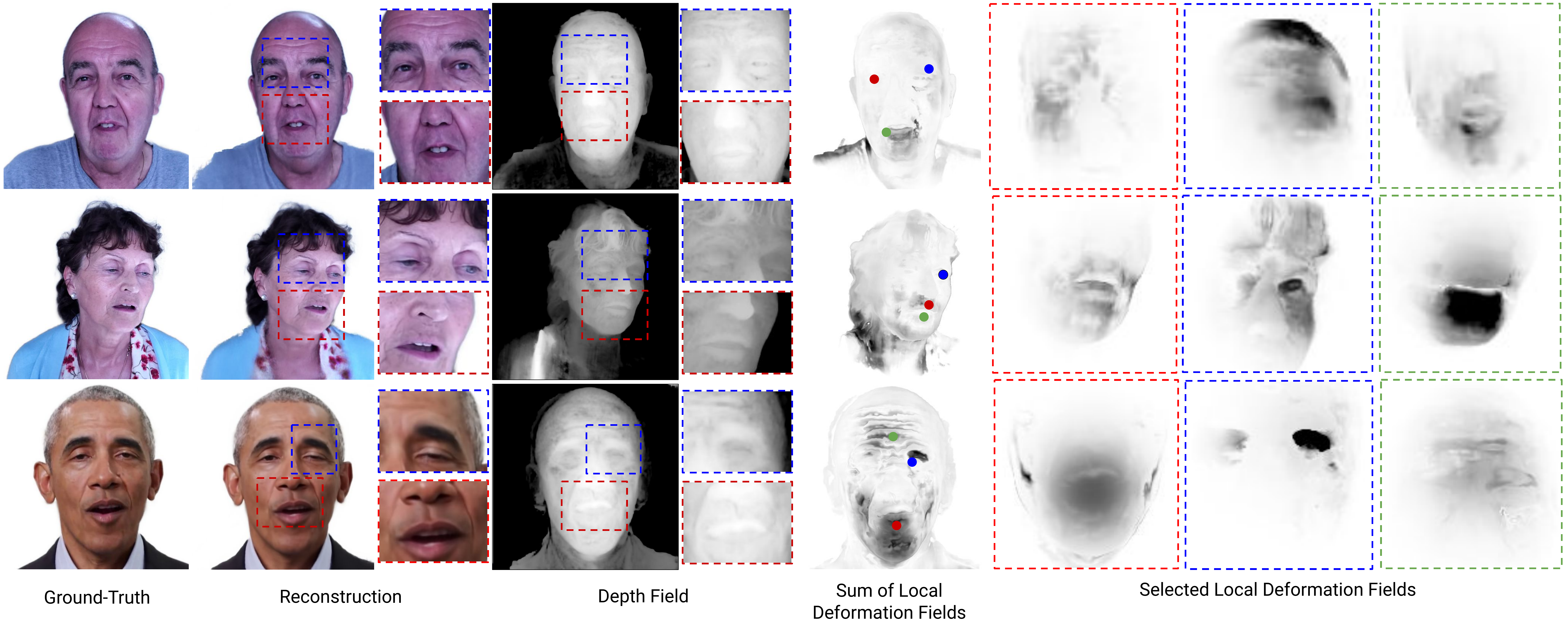}\vspace{-5pt}
    \caption{\textbf{Main results on test samples.} Our method models dynamic 3D deformations as an ensemble of local deformation fields, centered around 3D facial landmarks, shown as red, blue, and green dots in this example (details in~\cref{sec:method}). Our formulation synthesizes 3D neural heads from 2D videos with fine-grained geometric details, as shown in column 3 (depth field).}\vspace{-10pt}
    \label{fig:results}
\end{figure*}

In this paper, we propose a novel approach that allows implicit 3D rig modeling, and local control of detailed facial deformations. We use expression parameters obtained from a 3DMM face tracker, \eg, DECA~\cite{feng2021_deca}, but learn to model more \emph{local} deformations beyond linear 3DMM. To this end, instead of learning a single global deformation field, we break it into multiple local fields with varying spatial support, triggered by sparse facial landmarks, with weak supervision on the 3DMM expressions. The local deformation fields are represented by nonlinear function within a certain radius, and conditioned by tracked expression parameters weighted by an attention mask that filters redundant parameters that do not influence the landmarks. Finally, the local deformations are summed with distance-based weights, which are used to deform the global point to the canonical space, and retrieve radiance and density for volumetric rendering.

While part-based field decomposition~\cite{siarohin2023unsupervised,weng2022humannerf} approaches have been proposed, we demonstrate that decomposing implicit deformation fields into local fields improves representation capacity to model facial details. By filtering redundant input expression parameters to each local field and providing weak supervision via 3DMM, we achieve better, detailed local control, and modelling of asymmetric expressions (see~\cref{fig:results}). We provide qualitative and quantitative comparisons with state-of-the-art monocular head avatar synthesis methods and show that our approach reconstructs facial details more accurately, while improving local control of the face. In summary, our contributions are as follows:
\begin{enumerate}
\item A novel formulation that models local field deformation for implicit NeRF face rigs that provides fine-scale control with landmarks.\vspace{-5pt}
\item Local deformation fields surpass linear 3DMM's representation capacity via a novel local field control loss, and attention masks filtering.\vspace{-5pt}
\item We demonstrate the advantage of our approach in different applications, including detailed asymmetric expression control, and high-quality head synthesis with sharp mouth interior.\vspace{-5pt}

\end{enumerate}

\section{Related Work}
\label{sec:relatedwork}
The literature on detailed dynamic human avatars is vast. Thus, we mainly review methods for reconstructing controllable head models from unstructured 2D data using 3D aware neural rendering approaches. Please refer to Egger~\etal~\cite{egger2020_3dmfm} for a comprehensive overview on 3DMM and applications and to Tewari~\etal~\cite{tewari2020_sota-neural-render,tewari2022_advances-neural-render} for advances in neural rendering.\vspace{-6pt}

\vspace{-5px}
\paragraph{Neural Avatar Reconstruction and Portrait Synthesis}
Advances in neural rendering have pushed the boundaries of human avatar digitization from unconstraint 2D images or video. An early trend are model-based neural reconstruction approaches that learn to regress detailed 3DMMs \cite{shamai2019_synthesizing-facial, tran2019_towards,tewari2019_fml, br2021_learn-complete} or detailed rendering layers of a 3D model \cite{tewari2018_self-supervised,yamaguchi2018_high-fidelity,nagano2018_pagan,grecer2019_ganfit,lattas2022_avatarme++} using autoencoders or GANs to reconstruct high-quality 3D faces with photo-realistic appearance. However, model-based approaches often generate avatars with coarse local control derived from 3DMMs and restricted to the inner face region, \ie, they lack eyes, mouth interior, ears, and hair.
Another line of research attempts to synthesize face portraits using controllable GAN-based generative approaches \cite{kim2018_dvp, zakharov19_fs-talking-head, meshry2021_learned-spatial, wang2021_one-shot, zhou2021_pose-controllable}, driven either via sparse keypoints \cite{zakharov19_fs-talking-head, wang2021_one-shot,meshry2021_learned-spatial}, dense mesh priors \cite{kim2018_dvp, chandran2021_rendering-style, tewari2020_pie}, or multi-modal input \cite{wen2020_photoreal, zhou2021_pose-controllable}. These methods can synthesize photo-realistic 2D face portraits, but struggle with large poses and cannot generate detailed localized deformations, \eg, around mouth and eyes. The work by Lombardi~\etal~\cite{lombardi2018_deep-appear} and Ma~\etal~\cite{ma2021_pixel-codec} can generate 3D avatars with very detailed facial features, but local facial control is non-intuitive and reconstructions require expensive multiview camera systems.
Our approach reconstructs 3D avatars with detailed local control from unconstrained 2D videos.\vspace{-5pt}

\vspace{-10px}
\paragraph{Deformable Neural 3D Consistent Representations}
Modeling 3D-aware objects especially heads has been an active research in the last few years. Some approaches learn detailed 3D consistent implicit head models from a large corpus of unconstrained 2D images via 3D neural implicit surfaces \cite{ramon2021_h3dnet, or-el2022_stylesdf}, 3D-aware generative NeRFs \cite{deng2021_gram, chan2021_pigan, gu2022_stylenerf} or hybrid volumetric representations \cite{chan2022_efficient-geom-aware, wang2021_learn-compositional}.
NeRFs have particularly earned popularity since they can reconstruct complex scene structures while preserving fine scale details. Some recent NeRF-based approaches can model general dynamic objects, such as bodies and heads in general video scenes using dynamic embeddings and often by warping observed points into a canonical frame configuration \cite{park2021_hypernerf, pumarola2021_dnerf, park2021_nerfies,tretschk2021_nr-nerf, li2022_neural-3d-video}.
These approaches can reconstruct fine grained avatars from novel viewpoints but offer no semantic control over the generated models. Our approach on the contrary can learn detailed 3D consistent representations with local facial control.\vspace{-6pt}
\vspace{-10px}
\paragraph{Controllable Neural Implicit Head Generation}
Although implicit deformable head models have showed great promise, most methods lack semantic control over the learned implicit fields. A recent line of work create animatable implicit blendshape rigs with semantic latent representations such as landmarks and expression parameters to enable downstream animation and editing tasks \cite{athar2022_rignerf, zheng2021_imavatar, zheng2022_imface, sun2022_controllable, hong2021_headnerf, grassal2021_neural-head, gafni2021_dynamic-nerf, gao2022_nerfblendshape, cao2022_authentic}.
Some approaches learn deformation fields using dense mesh priors \cite{zheng2021_imavatar, athar2022_rignerf, gao2022_nerfblendshape}, sparse landmark priors \cite{zheng2022_imface} or hard mesh constraints \cite{sun2022_controllable,grassal2021_neural-head,ren2022_facial-geom}. Among them, \cite{zheng2021_imavatar, athar2022_rignerf} model pose aware deformation fields to enable full head control with better generalization capabilities at the expense of smoother reconstructions and coarse local control. Finer local control is achieved using multiple local deformation fields, either rigged via sparse landmarks \cite{zheng2022_imface} or blendshape weights \cite{gao2022_nerfblendshape}. However, these methods can only reconstruct mid-scale dynamic details, \ie, no wrinkles or sharp mouth interior. Hong~\etal~\cite{hong2021_headnerf} boost rendering quality by integrating a coarse-to-fine 2D neural rendering module with NeRF-based rendering. However, none these approaches can generate fine-grained facial deformations with sparse local control.
As far as we are aware, only implicit models derived from multiview imagery \cite{wang2022_morf, cao2022_authentic} can produce crisp results. Cao~\etal~\cite{cao2022_authentic} partially achieve personalized fine-grained control by fine tuning a universal implicit morphable model on user-specific custom expressions. As in Gao~\etal~\cite{gao2022_nerfblendshape}, local control granularity is limited to in-distribution expressions. On the contrary, we propose an implicit rig representation that learns to disentangle detailed deformation fields from arbitrary expressions and head poses while achieving good generalization power.

\begin{figure*}[ht]
    \centering
    \vspace{-10pt}
    \includegraphics[width=0.95\linewidth]{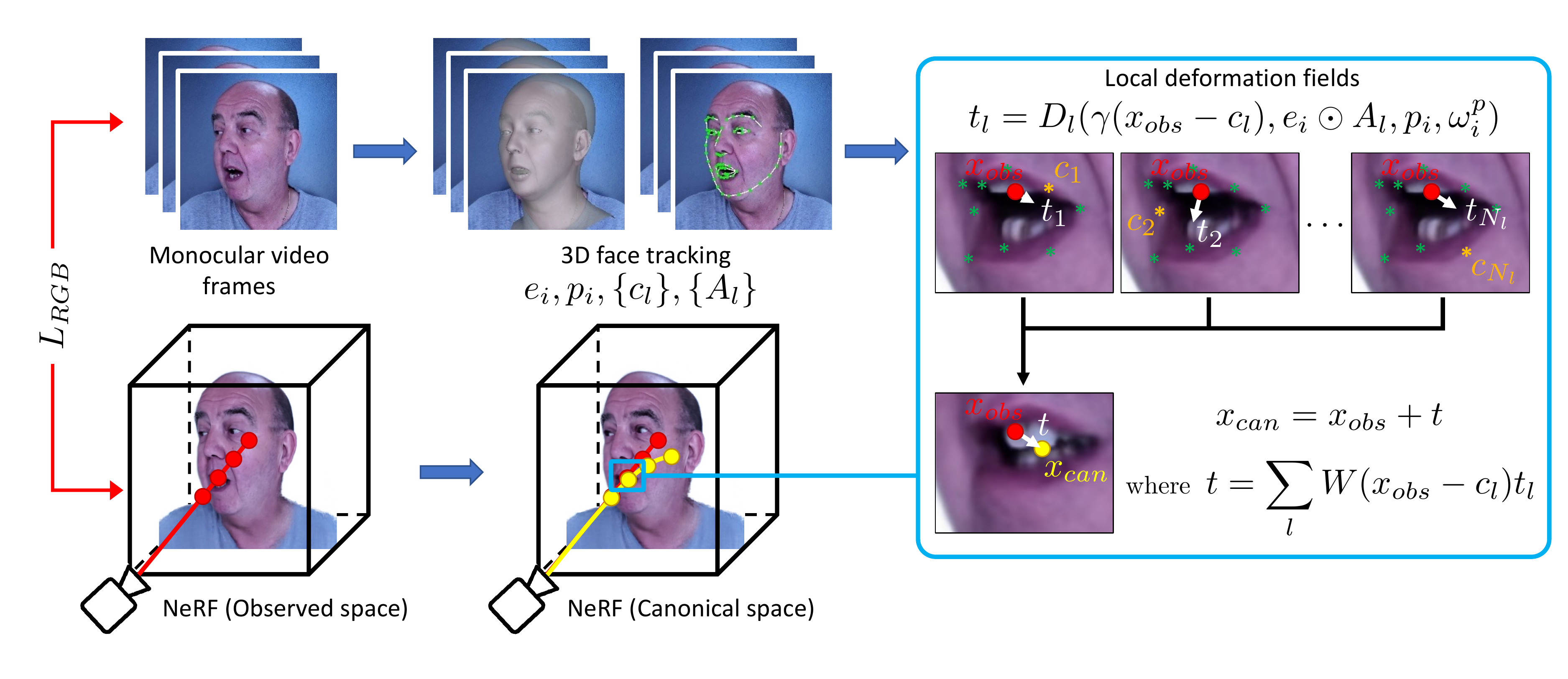}\vspace{-15pt}
    \caption{\textbf{Method overview.} Given an input video sequence, we run a face tracker~\cite{feng2021_deca} to get at each frame $i$ the following parameters of a linear 3DMM~\cite{li2017_flame}: expression $e_i$, pose $p_i$, and sparse 3D facial landmarks $c_l$. An attention mask with local spatial support is also pre-computed from the 3DMM. We model dynamic deformations as a translation of each observed point $x_{obs}$ to the canonical space, represented as $x_{can}$ (\cref{sec:method-dynamic_face_nerf}). We decompose the global deformation field $t$ into multiple local fields $\{t_l\}$, each centered around representative landmarks ${c_l}$ (\cref{sec:method-local_deformation_fields}). We enforce sparsity of each field $t_l$ via an attention mask $A_l$ that modulates $e_i$ (\cref{sec:method-attention_mask}). Our implicit representation is learned using RGB information, geometric regularization and priors, and a novel local control loss (\cref{sec:method-losses}).}\vspace{-5pt}
    \label{fig:method_overview}
\end{figure*}

\section{Our Method}
\label{sec:method}
In this section, we describe our method that enables detailed reconstruction of facial expressions with the ability to perform local control, as shown in~\cref{fig:method_overview}. We propose a deformable NeRF approach conditioned on expression parameters from a 3D face tracker, which learns translation from each point in the observed space to the canonical space, as defined by a 3DMM, \eg, FLAME \cite{li2017_flame}. To improve the ability to represent high-frequency change in the deformation field and allow local control, we break down the deformation field into multiple local fields conditioned on expression filtered by an attention mask. The locality is further enforced through a local control loss for each local field and the sum of the deformations from all local fields are weakly supervised by a mesh prior.

\subsection{Deformable Neural Radiance Field for Faces}
\label{sec:method-dynamic_face_nerf}
A neural radiance field is defined as a continuous function
\begin{eqnarray}
F: (x, d) \xrightarrow[]{}( \sigma(x), c(x,d))
\end{eqnarray}
that maps view direction $d$ and per-point location in the observed space $x$ into per-point density $\sigma$ and radiance $c$. Every pixel of the rendered scene is the accumulation of per-point density and radiance on each ray, cast through the pixel using the volumetric rendering equation \cite{mildenhall2022_nerf}.

To model dynamic scenes, we represent a continuous transformation from a point $x_{obs}$ in the observed space to the corresponding point $x_{can}$ in the canonical space \cite{park2021_nerfies} as
\begin{equation}
    T = D(\gamma(x_{obs}), \omega_i) \text{,}
\end{equation}
where $\omega_i$ is a per-frame deformation code that learns the scene at frame $i$, $\gamma$ are sinusoidal positional encodings of different frequencies \cite{tancik2020}, and $D$ represents a neural deformation field capable of modeling high-frequency variations.
A canonical point $x_{can}$ can then be obtained as follows:
\begin{eqnarray}
x_{can} = T(x_{obs}) \; .
\end{eqnarray}

Inspired by \cite{zheng2021_imavatar,athar2022_rignerf}, we model human face deformation in a monocular video setup instead of a per-frame deformation code. We condition the deformation field using a per-frame expression $e_i$ and pose $p_i$ code generated from an existing tracker as well as additional latent codes to account for pose inaccuracies $\omega^p_i$ and appearance inconsistencies $\omega^a_i$, so that 
\begin{eqnarray}
T & = & D(\gamma(x_{obs}), e_i, p_i, \omega^p_i) \\
\sigma, c & = & F(x_{can}, d, \omega^a_i)
\end{eqnarray}
Here, $T$ can be in $SE(3)$ to represent a 3D rigid deformation \cite{park2021_nerfies, park2021_hypernerf} or in $\mathbb{R}^3$ to model a non-rigid deformation \cite{pumarola2021_dnerf,li2021_neural-scene}. Since facial expression deformations are commonly non-rigid, we predict $T$ as translation $t\in\mathbb{R}^3$, and thus map $x_{obs}$ into $x_{can}$ as follows:
\begin{eqnarray}
x_{can} = t+x_{obs} \; .
\end{eqnarray}
While this disentanglement of deformation and canonical space provides better generalization to in- and out-of-distribution deformations, the additive model still struggles to represent details such as deep skin creases, crowfeet around eyes, and facial hair, especially when no ground truth mesh is available. The main reasons are that (i) a global model often lack expressibility \cite{zheng2022_imface} and (ii) the conditioning with 3DMM expression parameters represent linear and non-sparse local deformations (\cref{sec:analysis}).

\subsection{Local Deformation Fields}
\label{sec:method-local_deformation_fields}
Inspired by advances in part-based implicit rigging \cite{zheng2022_imface, zheng2022_structured, noguchi2021_narf, peng2021_animatable}, we overcome limitations of previous work by decomposing the global deformation field into multiple local fields, each centered around a pre-defined facial landmark location, to model non-linear local expression deformations with higher level of details, as shown in~\cref{fig:method_overview}.

Mathematically, we first define a set of $N_l$ landmarks on the face mesh and a set of local deformation fields $D_l$, $\forall l \in [1,...,N_l]$. 
We denote $c_l$ to be the 3D position of the $l^{\text{th}}$ landmark on the deformed mesh. 
To map a point $x_{obs}$ from the observed space to its corresponding point in the canonical space, we first compute the position of this point relative to a landmark location (\ie, $x_l = x_{obs} - c_l$) and pass this to the local deformation field $D_l(\gamma(x_l))$.  We repeat this process for all landmarks, and use a weighted average to compute the deformation:
\begin{eqnarray}
t &=& \sum_{l}W(x_l)t_l \; ,\\
\text{where} \quad \; t_l &=& D_l(\gamma(x_{l}), e_i, p_i, \omega^p_i) \; .\\ \nonumber
\end{eqnarray}
Similar to \cite{zheng2022_structured}, we define the weight $W(x)$ as a scaled Gaussian function with zero mean and $R$ standard deviation based on the point's distance to each local center, \ie,
\begin{equation}
W(x) = \displaystyle \max\left( \exp{\left(\frac{-\|x\|_2^2}{2 R^2}\right)} - \tau, 0\right) s \; ,
\end{equation}
where $R$ is a user-defined parameter representing the spatial support of $t_l$, $\forall l$. The threshold $\tau$ prevents points farther away from each landmark to query the local neural network, improving computing efficiency. The scale $s < 1$ adjusts the magnitude of the local fields' outputs to the range of valid values for facial deformation.

\subsection{Attention Mask Based on 3DMM Bases}
\label{sec:method-attention_mask}
Although the expression bases offer global spatial support, some parameters have little effect on certain local regions. As such, feeding the same expression code into each local field is redundant. 
We can prune these parameters by leveraging an attention mask with local spatial support to help enforce sparsity of each local field, \ie, 
\begin{eqnarray}
t_l = D_l(\gamma(x_{l}), e_i \circ A_l, p_i, \omega^p_i) \; ,
\end{eqnarray}
where $A_l$ $\in [0,1]^{|e|}$ is a binary attention mask associated with the expression parameters $e$ of each local deformation $D_l$ and $\circ$ is the element-wise product.

We precompute $A_l$, $\forall l$ in a one-time preprocessing step by exploiting the spatial support of expression bases at each landmark location $l$ of the 3DMM.
We first compute a distance matrix $\mathcal{D}$ $\in \mathbb{R}^{N_l \times |e|}$ containing absolute delta displacements of each expression basis activated independently over all landmark locations. 
Each column $\mathcal{D}_{*,i}$ represents the spatial support of the $i$-th expression basis over all $N_l$ landmarks.
For each column $\mathcal{D}_{*,i}$ we zero out rows where the spatial support of landmark locations fall below the $25\%$ quantile. We then binarize $\mathcal{D}$, yielding a binary cross attention mask $A$.\looseness=-1

\begin{figure*}[ht]
 \includegraphics[width=\textwidth]{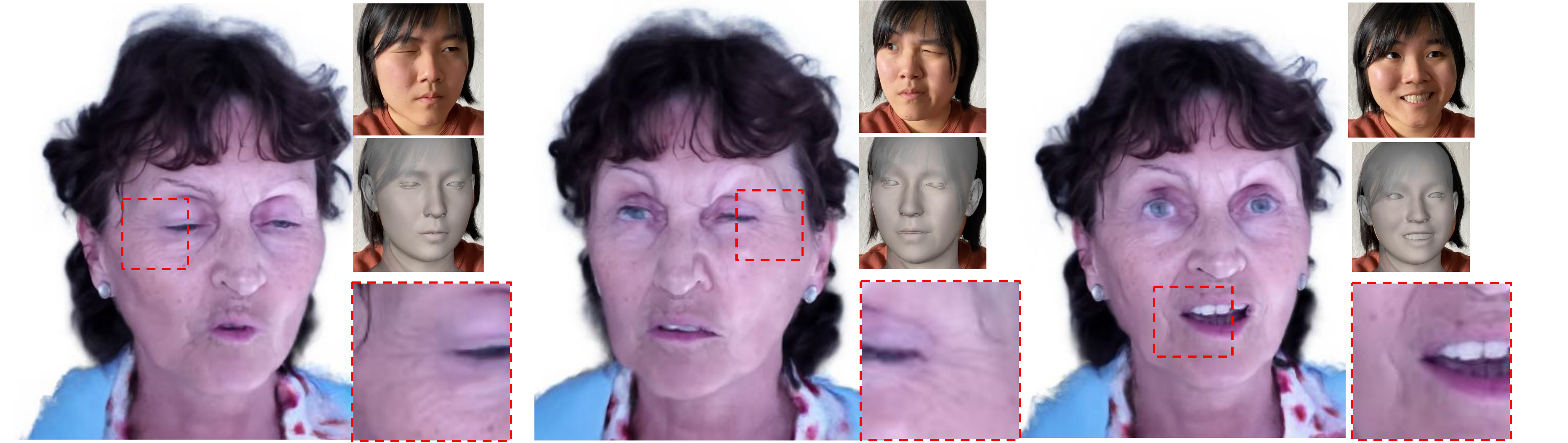} 
\caption{\textbf{Novel pose and expression synthesis via face reenactment}: We track the user's expressions (top) via DECA and transfer the 3DMM parameters (middle) to the neural head model of \emph{Subject2}. The model produces asymmetric expressions under the user's pose with a high level of details (bottom) that surpasses the linear 3DMM. Note that none of the transferred expressions were in the training set of \emph{Subject2}.\looseness=-1}\vspace{-8pt}
\label{fig:local_control}
\end{figure*}

\subsection{Training Objectives}
\label{sec:method-losses}
\paragraph{RGB Loss} Our method trains in a self-supervised manner by comparing a rendered and ground truth RGB pixel
\begin{eqnarray}
L_{RGB} = \frac{1}{|P|} \sum_p\|\hat{I}_p-I_p\|_2 \; ,
\vspace{-20px}
\end{eqnarray}

where $P$ is the set of sampled training pixels, and $\hat{I}_p$ and $I_p$ denote rendered and measured pixel values, respectively. 
We use an $\ell_{2,1}$-norm over the sampled pixel locations, akin to \cite{tewari2018_self-supervised}, to allow sharper per-pixel reconstruction while maintaining the same convergence speed.\vspace{-10px}

\paragraph{Local Control Loss}
We propose a novel loss function to enforce locality of each local field. The 3D facial landmarks share the same semantic face location for all poses and expressions. Therefore, the same 3D facial landmark across all frames should map to the same position in the canonical space.  Based on this insight, we enforce that all rays cast toward a facial landmark in the observed space intersect the face geometry at the same point in the canonical space.

Mathematically, given any two frames and corresponding latent codes $\delta_1=(p_1,e_1)$ and $\delta_2=(p_2,e_2)$, for each keypoint $c_l$, we then cast a ray based on the pose of each of the 2 images to the deformed keypoint location $c_{l_1}$ and $c_{l_2}$.  

Next, we solve a volume rendering integral to compute depth associated with the rays, by  (i) sampling points along each ray, (ii) querying the local deformation field at these points, and (iii) and evaluating the density at the corresponding points in the canonical space. We take the expected value of the densities to get depth, which determines the 3D points $\hat{x}_{1,l}$ and   $\hat{x}_{2,l}$ in deformed space where the rays and face geometry intersect. (iv) We then deform the intersection points to the canonical space and enforce that the estimated surface intersection should be the same in the canonical space, achieving semantic local control implicitly via intersection equivariance constraints, with respect to the $\ell_{1}$-norm as follows:
\vspace{-10px}
\begin{eqnarray}
L_{local} = \sum_l \|\hat{x}_{1,l} -  \hat{x}_{2,l}\|_1
\end{eqnarray}
This loss helps increase sparsity of the local deformation field, hence allowing for better locality of expression control, as well as enabling sharper and more consistent local deformations, as shown in~\cref{sec:analysis}. We also observed that using $L_1$ loss is sufficient to remove effect of the outliers from occluded 2D keypoints.

\paragraph{Weak Mesh Prior}
We also leverage prior geometric knowledge from the tracked 3D mesh. Specifically, for each sampled point $x$ along the ray we first query the closest surface point $x_m$ on the deformed mesh $M$ to get a pseudo ground truth mesh constraints.

We then supervise the predicted deformation $t(x)$ with pseudo ground truth labels for salient face regions as follows:
\begin{eqnarray}
L_{mesh} = \frac{1}{|X|} \sum_x F_{surf}(x) \cdot \|t(x) -t_{mesh}(x_m) \|_2 \; ,
\end{eqnarray}
where $X$ is the set of sampled training points and $t_{mesh}(x_m)$ is the pseudo ground truth deformation obtained by first finding the vertex deformation of the 3 vertices of the triangle containing $x_m$ and then interpolating at $x_m$ via barycentric coordinates. $F_{surf}(x)$ is a binary mask that contains non-zero entries for sampled points $x$ that (i) represent foreground pixels as defined by a segmentation mask and (ii) are visible surface face regions, \ie, rendering weight (per-point alpha * transmittance) is above a user-defined threshold $\tau = 10^{-4}$.\vspace{-5pt}

\paragraph{Deformation Field Regularization}
We penalize large deformations of the reconstructed field $t$ by enforcing $\ell_2$-norm regularization of per-point deformations
\begin{eqnarray}
L_{def} = \frac{1}{|X|} \sum_x \emph{lut}(x) \cdot \|t(x)\|_2 \; ,
\vspace{-10px}
\end{eqnarray}
where $\emph{lut}(x)$ is a look-up table that assigns a higher weight to points sampled from background pixels.\vspace{-5pt}
\paragraph{Volume and Latent Codes Regularization}
Following the work in \cite{chen2022_tensorf}, we impose $\ell_1$ sparsity loss $L_{vol}$ on volume density. Furthermore, we penalize large latent code variations $\omega = ( \omega^p , \omega^a )$ to avoid per-frame overfitting. Specifically, we use $L_{code} = \|\omega\|_2$, akin to \cite{gafni2021_dynamic-nerf,park2021_nerfies}.

Overall, our final loss is represented as follows:
\begin{eqnarray}
L & = & L_{RGB} + \lambda_{local} L_{local} +  \lambda_{mesh} L_{mesh} + \nonumber \\
& & \lambda_{def} L_{def} + \lambda_{vol} L_{vol} + \lambda_{code} L_{code}
\end{eqnarray}

\subsection{Implementation Details}
\label{sec:implementation}

Our framework models the canonical space with TensoRF \cite{chen2022_tensorf}. We use Adam optimizer \cite{kingma2014_adam} with a learning rate of $1e^{-3}$ decayed to $1e^{-4}$ at end of training. We pre-train on each sequence without deformation field and latent codes for 14k iterations to learn a rough estimate of the canonical space. We use $N_l=34, R=0.03, \tau=1e-4, s=0.02$, and 10 positional encoding frequencies linear in log space for each local field and we increase the weight on positional encoding frequencies \cite{park2021_nerfies} over 40k iterations. We utilize a ray batch size of $2048$ and train for 400k iterations. The following weights are used in our training objective: $\lambda_{local} = 0.01$, $\lambda_{mesh}=0.01$, $\lambda_{field}=0.5$, $\lambda_{code}=0.01$, and $\lambda_{vol}=2e^{-4}$ at the start of training and then we change it to $6e^{-4}$ after the first upsampling of the voxel grid. We use $\emph{lut}(x)=\{1.0,100.0\}$ for penalizing foreground and background sampled points, respectively. The dimension of $w^p$ and $w^a$ is 32 and 16, respectively. The final deformation from each local field is the sum of the output of an MLP and the deformation of the centroid $c_l$. All the hyperparameter values were found empirically and used in all our experiments.\looseness=-1
We apply the tracker DECA \cite{feng2021_deca} to compute expression and pose parameters, 3D landmarks, and tracked meshes for each sequence. This data is used for conditioning our deformation fields, finding the deformation field centroids and intersection points, and weakly supervising deformation fields, respectively. To run TensoRF with tracked results from DECA, we first obtain the global translation of the head from the camera described in \cite{zheng2021_imavatar} and  convert the product of global translation and rigid head pose $[R|t]$ into camera extrinsics.
We generate the face segmentation mask with FaceParsing \cite{zheng2022_farl} and mask out the background with alpha matte from MODnet \cite{ke2022_modnet}.

\section{Results}
\label{sec:results}

In this section, we show the effectiveness of our local deformation fields in synthesizing novel expressions and poses on real videos. 
We conduct quantitative and qualitative comparisons with three state-of-the art baselines on implicit neural synthesis, and show a facial reenactment application for unseen asymmetric expressions.

\subsection{Datasets}
\begin{figure}[t!]
\centering
 \includegraphics[width=\linewidth]{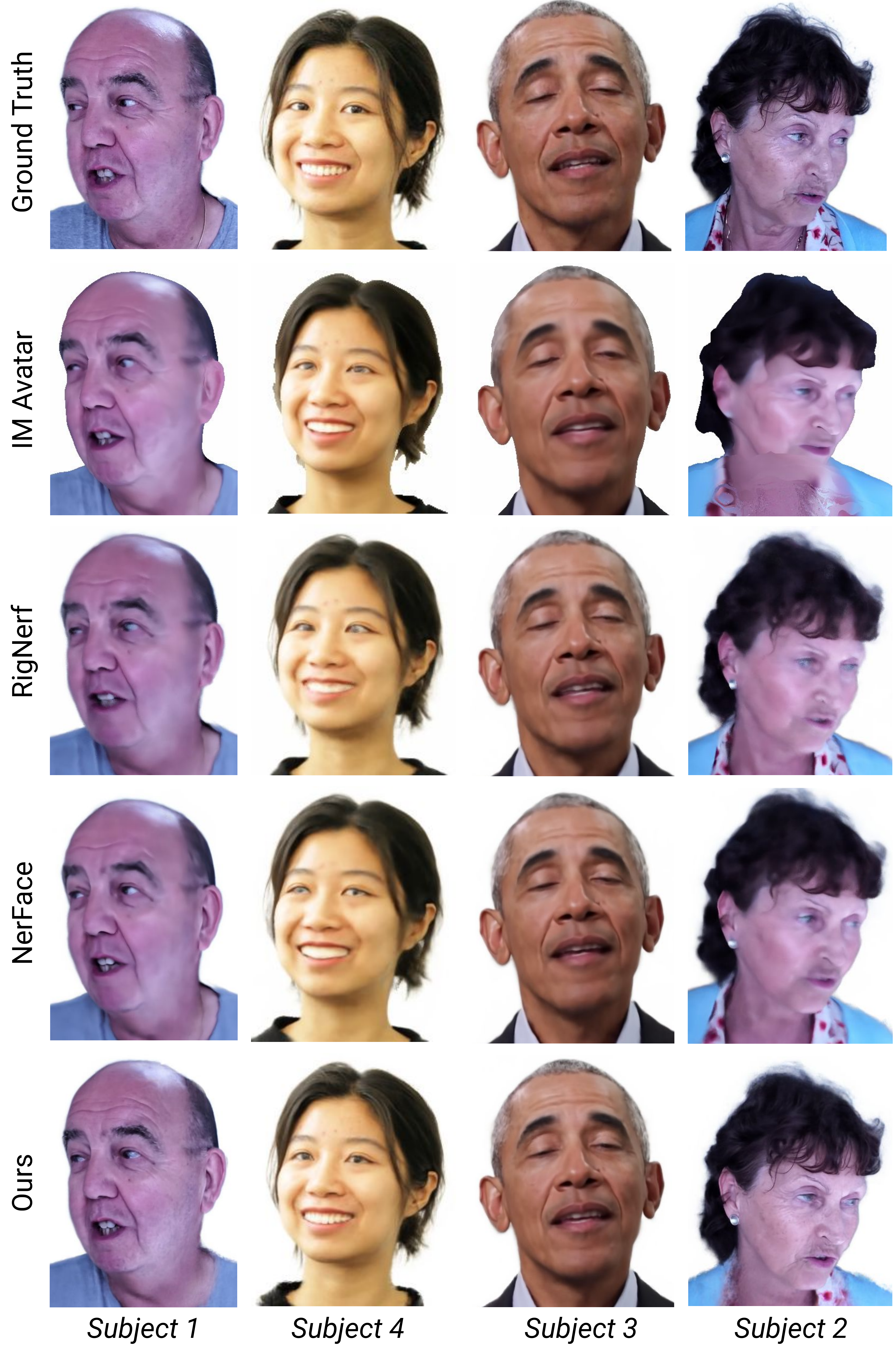}
\caption{\textbf{Qualitative comparisons with state-of-the-art on test data.} Top to bottom: GT, IM Avatar(-), RigNeRF*, NerFace, and our approach. Overall our approach synthesizes fine-scale skin details e.g wrinkles and moles, and facial attributes at higher fidelity. Note that IM Avatar(-) is a version that synthesizes sharper details at the expense of poorer pose and expression extrapolation.}\vspace{-5pt}
\label{fig:comparison}
\end{figure}

We evaluate our method on monocular video sequences of four real subjects, as shown in~\cref{fig:comparison}.
We divide each video sequence into 6.5k images for training and 1.5k--2k images for testing, except for \emph{Subject4} sequence, which is split into 2.7k and 1.8k training and testing images, respectively.
We crop all frames around the face center and run our method at $500\times500$ resolution, except for \emph{Subject4} sequence, which we run at $256\times256$ to match IM Avatar's resolution~\cite{zheng2021_imavatar}.
Please refer to the supplementary material for a detailed description of our dataset as well as a visualization of pose and expression distribution.

\subsection{Pose and Expression Synthesis and Control}
\cref{fig:results} shows our reconstruction results on test sequences of \emph{Subject1} and  \emph{Subject2}. Overall, our approach can render heads with detailed mouth interior, and eyes and skin that look almost indistinguishable from the ground truth. As seen in column 3, the depth field reveals very detailed geometry, especially for \emph{Subject2}, which has not been demonstrated before by previous approaches as shown in the supplementary material. In addition, we show that for \emph{Subject3} we can perform local expression control by injecting a different expression code to a particular local field, in this case the one around the right eye, allowing the right eye to wink.

\cref{fig:local_control} illustrates reenactment results for \emph{Subject2} to demonstrate that our approach renders high-quality avatars with localized expression control and crisp details in the eyes and mouth interior, even for highly non-symmetrical expressions, which are also out-of-distribution.
Note that \emph{Subject2} especially column 2 can reproduce non-linear deformations as performed by the user that go beyond the representation power of the underlying 3DMM.

\subsection{Qualitative Comparisons}
We compare our approach with three implicit neural head synthesis baselines: IM Avatar(-)\footnote{IM Avatar (-) refers to the version trained without FLAME supervision.}~\cite{zheng2021_imavatar}, NerFace~\cite{gafni2021_dynamic-nerf}, and RigNeRF*\footnote{We reimplemented RigNeRF. See details in \cref{sec:supp_implementation}.}~\cite{athar2022_rignerf}. We train the baseline approaches with default settings on our training set and evaluate on every 10th frame of the test set. Since in RigNeRF* and our approach, the deformation field is conditioned on a per-frame deformation latent code, to evaluate against the testset, we first optimize for the deformation latent code on each test frame with RGB loss while fixing all other parameters for our model for 10k iterations and RigNeRF* for 40k iterations. For NerFace and IM Avatar, we use the latent code of the first training frame. For reenactment, we use both the deformation latent code and appearance latent code of the first training frame. As discussed in \cite{gafni2021_dynamic-nerf}, per-frame latent codes are used to adjust for small inaccuracies in pose and expressions for better convergence. See  supplementary material, \cref{tab:comparisons_real-extended} and
\cref{fig:comparisons_normal}, for comparisons without test-time latent code optimization.
\cref{fig:comparison} shows qualitative comparisons with state-of-the-art approaches on our test dataset. Our approach synthesizes fine-scale skin details at higher fidelity and overall attains sharper reconstructions for the eyes, mouth interior, and hair than previous approaches. Note that wrinkles, moles and teeth are reproduced by our approach at a level of detail not seen before. 
On \emph{Subject4} sequence, state-of-the-art methods are on par with our approach, stemming from the low image resolution of the dataset.

In our experiments, we observed that IM Avatar full model, \ie, the model with FLAME supervision, produced overly smoothed renderings, making comparisons unfair. For this reason, we trained IM Avatar(-) with RGB and mask loss only to boost sharpness. While it may result in worse ability to generalization to novel pose and expression, it is not the purpose of our test.
\subsection{Quantitative Comparisons}

\label{sec:quantitative}

To assess the quality of the rendered avatars, we adopt the metrics by Gafni \etal~\cite{gafni2021_dynamic-nerf}. Specifically, we measure the Manhattan distance ($\ell_1$) in the RGB color space, PSNR, LPIPS \cite{zhang2018_unreasonable-effectiveness}, and SSIM \cite{wang2004_image-quality}.
\cref{tab:comparisons_real} shows that our approach generates images with higher fidelity on \emph{Subject1}, \emph{Subject2}, and \emph{Subject3}, and that outperforms state-of-the-art methods on all sequences based on LPIPS metric.
However, $\ell_1$, SSIM and PSNR metrics degrade on \emph{Subject4}. This dataset is particularly challenging as test samples are mainly out of distribution. Unlike IM Avatar(-) and RigNeRF*, our approach trades quality more than robustness to unseen data. Still, our approach generates perceptually better renderings.

\begin{table*}
\footnotesize

	\begin{center}
\resizebox{1\linewidth}{!}{
\begin{tabular}{l|llll|llll|llll|llll}
          & \multicolumn{4}{c|}{\emph{Subject1}}                                           & \multicolumn{4}{c|}{\emph{Subject2}}                                             & \multicolumn{4}{c|}{\emph{Subject3}}                                        &    \multicolumn{4}{c}{\emph{Subject4\cite{zheng2021_imavatar}}}                                       \\
          & $\ell_1$ $\downarrow$            & SSIM  $\uparrow$           & LPIPS $\downarrow$         & PSNR $\uparrow$           & $\ell_1$ $\downarrow$             & SSIM $\uparrow$          & LPIPS$\downarrow$           & PSNR $\uparrow$            & $\ell_1$ $\downarrow$             & SSIM $\uparrow$          & LPIPS$\downarrow$          & PSNR $\uparrow$           & $\ell_1$ $\downarrow$ $\downarrow$            & SSIM $\uparrow$          & LPIPS $\downarrow$         & PSNR $\uparrow$           \\ \hline
Nerface   & 0.058          & 0.903          & 0.105          & 21.457          & 0.077          & 0.889          & 0.121           & 18.252           & 0.047          & 0.894          & 0.055          & 22.220 & 0.077          & 0.818 & 0.085          & 17.910                    \\
IM Avatar(-) & 0.068          & 0.901          & 0.113          & 20.502          & 0.093          & 0.877          & 0.157           & 14.960           & 0.043          & 0.900          & 0.078          & 23.218 & \textbf{0.063} & \textbf{0.870}          & 0.069          & 19.215                    \\
RigNeRF*   & \textbf{0.055}          & 0.904          & 0.095          & 22.324          & 0.072          & 0.884          & 0.120           & 18.922           & 0.035 & 0.910 & 0.052          & 24.634 & 0.065          & 0.844          & 0.063          & \textbf{19.253}  \\ \hline
Ours      & 0.056 & \textbf{0.934} & \textbf{0.0465} & \textbf{23.656} & \textbf{0.062} & \textbf{0.917} & \textbf{0.0576} & \textbf{20.4375} & \textbf{0.0206}          & \textbf{0.971}          & \textbf{0.0265} & \textbf{30.854} & 0.081          & 0.830          & \textbf{0.062} & 19.085          
\end{tabular}
}\vspace{-13pt}

    \end{center}
    \caption{\textbf{Quantitative comparisons with state-of-the-art on test data.} Our approach achieves state of the art performance using LPIPS metric and generates better metrics on \emph{Subject1}, \emph{Subject2}, and \emph{Subject3}. RigNeRF* and especially IM Avatar(-) generalize better to \emph{Subject4}.}\vspace{-5pt}
    \label{tab:comparisons_real}
\end{table*}

\section{Analysis}
\label{sec:analysis}
We ablate on the use of attention mask, local control loss, and local fields vs global field and demonstrate qualitatively and quantitatively that each component assists in rendering unseen asymmetric expressions and reconstructing details.

\begin{figure}
\centering
 \includegraphics[width=\linewidth]{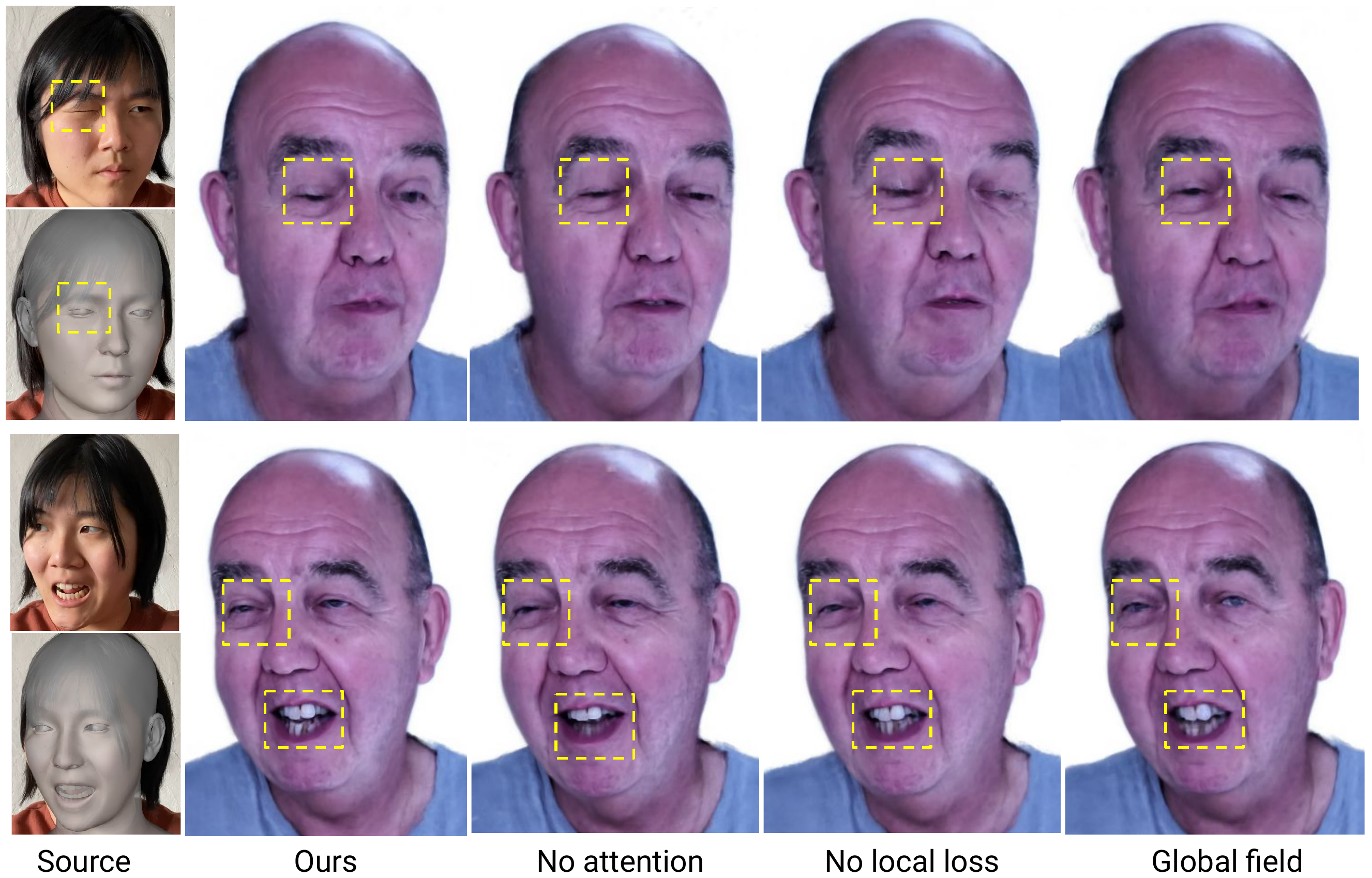}
\caption{\textbf{Ablation tests on \emph{Subject1} sequence.} Expression and pose reenactment with our full model, our model without attention mask, without Local Control Loss, and using global deformation field instead of local deformation fields.}
\label{fig:ablation}\vspace{-5pt}
\end{figure}

\vspace{-10px}
\paragraph{Influence of Attention Mask}
Attention masks help prune the effect from expression bases that have small influence on a local region and enforce sparsity of the local fields while retaining the representation power needed for face regions with large deformation such as the mouth. \cref{fig:ablation}, column 2 shows that without the attention mask, multiple local fields could be learned to contribute to the deformation caused by eye opening and closing. Since the training set only includes expressions where two eyes open and close at the same time, the influence of these local fields could not be fully disentangled, especially due to the global control of the expression codes.\looseness=-1

\vspace{-10px}
\paragraph{Local Control Loss}
Local control loss further improves locality by encouraging each local field to produce consistent and independent influence over different pose and expressions. As demonstrated by comparing \cref{fig:ablation} columns 1 and 3, adding local control loss allows independent control of local fields, while also improves convergence of the deformation fields which allows more accurate modeling of details such as the teeth.

\vspace{-10px}
\paragraph{Global vs Local Fields}
We also examine the influence of using multiple local fields instead of a global field. We represent the deformation field with an MLP with roughly the same number of parameters as the sum of parameters from all local fields, $10$ number of positional encoding frequencies and supervised with mesh prior on top of RGB loss. \cref{fig:ablation}, column 4 shows that while a global field with high frequencies can represent some details, it does not have sufficient representation power to model complex deformations such as mouth opening, resulting in defective teeth details. It also cannot model unseen asymmetric expressions beyond the global 3DMM bases. We also evaluated each experiment against the ground truth images on the test set for \emph{Subject1}, and computed metrics described in~\cref{sec:quantitative}, listed in~\cref{tab:ablation}. Using all of the above mentioned components improves the reconstruction quality quantitatively. 

\begin{table}
    \footnotesize
	\begin{center}
\resizebox{1\linewidth}{!}{%
\begin{tabular}{l|llll}
Variants          & \multicolumn{4}{c}{Metrics}                                          \\
                  & $\ell_1$   $\downarrow$           & SSIM $\uparrow$          & LPIPS  $\downarrow$         & PSNR $\uparrow$           \\ \hline
Global Field      & 0.0656          & 0.930          & 0.0552          & 22.467          \\
No Attention mask & 0.0582          & 0.932          & 0.0487          & 23.309          \\
No Local          & 0.0577          & \textbf{0.934} & 0.0486          & 23.421          \\
Ours              & \textbf{0.0558} & \textbf{0.934} & \textbf{0.0465} & \textbf{23.656}
\end{tabular}
}\vspace{-13pt}
    \end{center}
    \caption{\textbf{Ablation tests on \emph{Subject1} sequence.} Image and perceptual metrics with respect to our full model, our model without attention mask, without Local Control Loss, and using global deformation field instead of local deformation fields.}
    \label{tab:ablation}
\end{table}

\begin{figure}
\centering
 \includegraphics[width=0.9\linewidth]{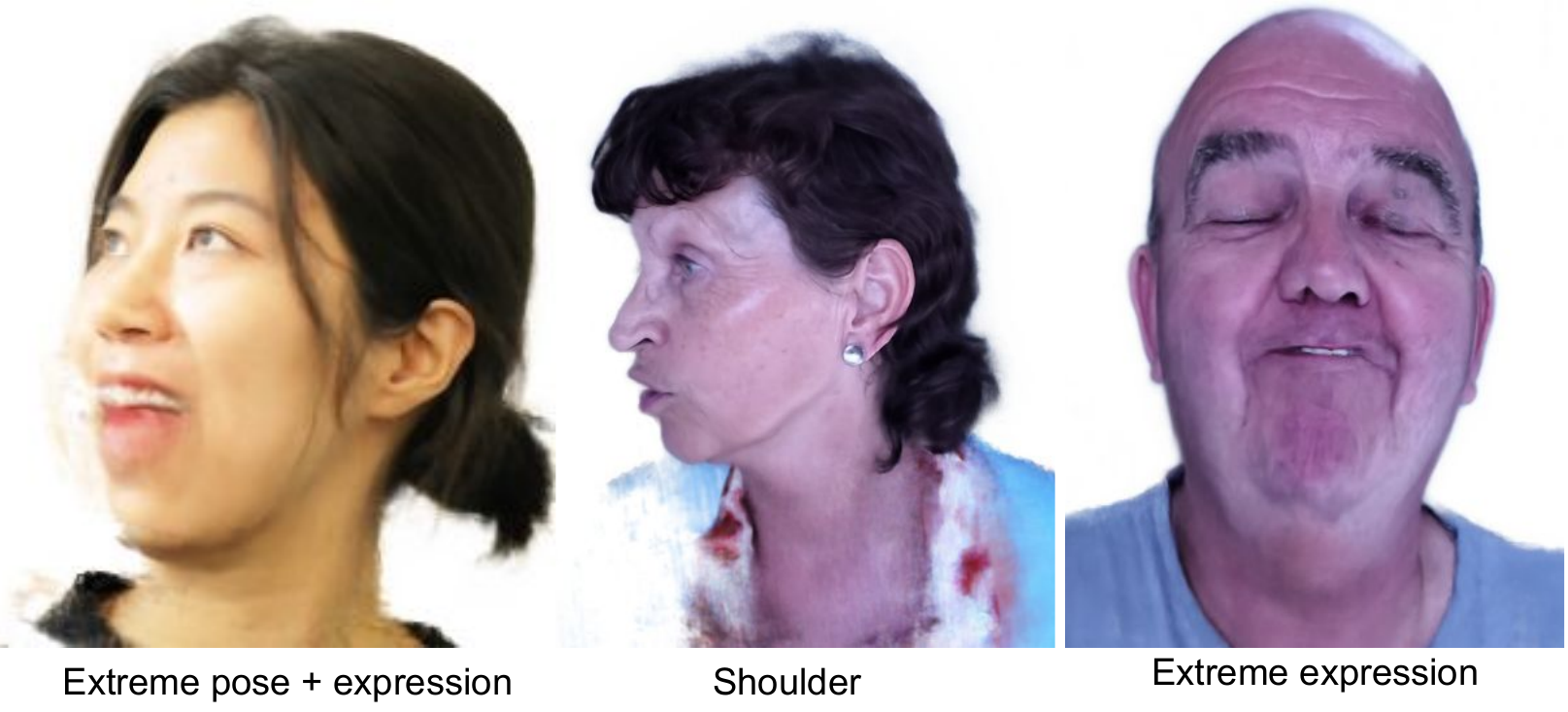}
\caption{\textbf{Limitations.} Defects with extreme poses and expressions, and shoulder when the dataset contains big shoulder movement.}\vspace{-5pt}
\label{fig:limitations}
\end{figure}

\vspace{-10px}
\paragraph{Limitations.} Our approach is designed to model fine-scale facial details. While we can handle a varying set of pose and expressions with fine details, the reconstruction quality degrades on extreme pose and expression variations~\cref{fig:limitations}. Moreover, pose and expression are not always fully disentangled in our \textit{in-the-wild} datasets. Thus, improved generalization due to extreme pose and expression variation is left as future work. Furthermore, non-facial parts of humans \eg shoulders, are currently not explicitly modeled in our formulation and are not exposed to local fields, leading to noisy and sometimes blurry torso renderings.
\section{Conclusion}
\label{sec:conclusion}
We present an approach to model face rigs with neural radiance fields that go beyond linear 3DMM and global deformations, and model nonlinear local deformation and fine-grained facial details. A novel local control loss helps enforce the locality and consistency of local deformation fields, which are controlled via a sparse landmark set and weakly supervised via 3DMM. Further, we introduce an attention mask to filter redundant expression parameters that have small influence on particular face regions, leading to more accurate local deformation fields, the subsequent expressions, and detail reconstruction. We attain state-of-the-art statistics on perceptual radiance image quality and demonstrated that our approach is able to reconstruct controllable local details. We demonstrate that it leads to several applications, such as detail preserving mouth interior articulation, asymmetric lips and cheek movement, and eye winking, among others. Additionally, we demonstrate the limitations of our approach to in-distribution poses and expressions, and improvements in generalization will be explored in the future.

\section{Acknowledgments}
We thank Mrs. and Mr. Mann for kindly accepting to be recorded and letting us show their recorded videos for this work, and anonymous reviewers for their valuable feedback.

{\small
\bibliographystyle{ieee_fullname}
\bibliography{main}
}

\clearpage
\appendix
\section{Implementation Details}
\label{sec:supp_implementation}

\paragraph{TensoRF}
We use TensorRF\cite{chen2022_tensorf} for the canonical space of our pipeline with the following architecture changes, we use 1) an appearance latent code into the MLP decoder to account for appearance inconsistencies, and 2) RELU activation to threshold volume densities instead of Softplus to allow sharper reconstruction. To combine the training of TensoRF and deformation fields, we freeze the deformation field and pre-train TensoRF for 14k iterations. During pre-training, we grow the voxel grid from $128^3$ to the maximum resolution $300^3$ (as in the original paper) for \emph{Subject1}, \emph{Subject2}, and \emph{Subject4} and $200^3$ for \emph{Subject3}. We also prune voxels with density smaller than $1e^{-4}$.

\begin{figure}[h]
\centering
 \includegraphics[width=0.8\linewidth]{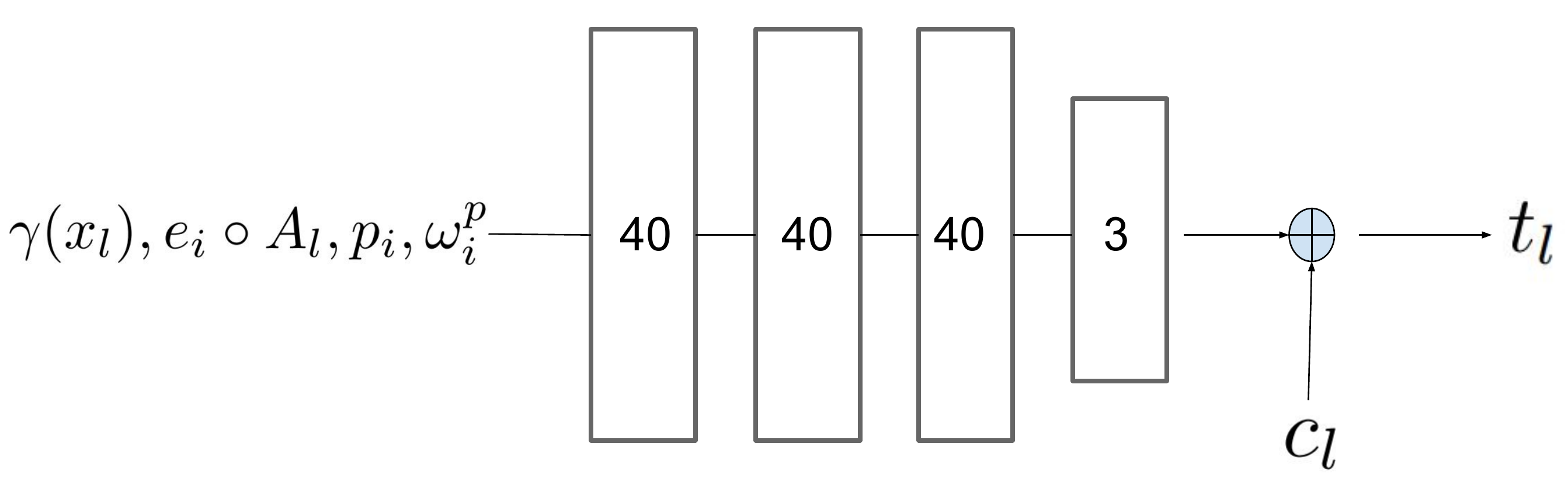}
\caption{\textbf{Architecture diagram of a single local field $D_l$}.}
\label{fig:arch}
\end{figure}

\paragraph{Deformation Field}
The architecture of the deformation field is shown in \cref{fig:arch}. Each local deformation field consists of a 3-layer MLP. Each layer consists of 40 neurons, followed by a Leaky RELU. The input to the input layer of each local field MLP is concatenation of a global expression code and jaw pose masked with the Attention Mask, global deformation latent code, and head and neck pose.

\paragraph{RigNeRF*}
We modify the original RigNerf~\cite{athar2022_rignerf} architecture for the monocular setting and refer to it as RigNeRF*, where the head pose from the tracker is transformed into a camera extrinsic matrix, as if the head remains static and the camera moves. Hence, for each frame, we query the mesh deformed by expression only, instead of both pose, and expression as in the original paper. We use the same network architecture, and parameters for the deformation field, as well as the same latent code dimensions.  Finally, we train each sequence for 1M iterations with ray batch-size of 1550 instead of the 100k epochs indicated in the original paper, as it is a less complex task for the deformation field to learn the deformation due to expressions only rather than both pose and expression.

\section{Dataset Analysis}

Our video dataset consists of four subjects, as shown in~\cref{fig:comparison}.  \emph{Subject1} (1st column) and \emph{Subject2} (4th column) are subjects captured indoors with a 4K phone camera.
\emph{Subject3} (3rd column) is ex-president Obama addressing a commencement speech, and is a segment of an HD video downloaded from YouTube.\footnote{Commencement speech of class 2020: \url{https://youtu.be/NGEvASSaPyg}} \emph{Subject4} (2nd column) is a female subject from IM Avatar benchmark dataset.\footnote{\url{https://dataset.ait.ethz.ch/downloads/IMavatar_data/data/yufeng.zip} We use sequence MVI\_1810 and MVI\_1814 as training set and MVI\_1812 as test set}
The first three datasets show unscripted natural expressions with varying head poses, while the latter is split into a speech video and another video with difficult expressions and poses, as described in \cite{zheng2021_imavatar}. \cref{tbl:supp_dataset_analysis} shows the neck, head, and jaw pose distribution of the four monocular sequences. Please refer to the supplementary video for detailed visualization of the poses as well as the range of expressions.

\begin{table}
\footnotesize
\resizebox{\linewidth}{!}{
\begin{tabular}{ll|lll|lll|lll}
                               &      & \multicolumn{3}{c|}{Neck} & \multicolumn{3}{c|}{Head} & \multicolumn{3}{c}{Jaw} \\ \hline
\multicolumn{1}{l|}{\emph{Subject1}} & MEAN & 6.09   & 0.314   & -1.14  & 1.53   & -6.44  & -0.105  & 7.37 & -0.0643 & 0.258  \\
\multicolumn{1}{l|}{}          & STD  & 6.75   & 4.21    & 6.02   & 10.1   & 15.10  & 6.63    & 1.89 & 2.26    & 3.24   \\ \hline
\multicolumn{1}{l|}{\emph{Subject2}} & MEAN & 1.27   & -0.432  & 0.143  & 3.63   & -1.69  & -2.67   & 3.94 & -1.02   & -1.26  \\
\multicolumn{1}{l|}{}          & STD  & 7.64   & 7.42    & 6.77   & 9.66   & 32.3   & 7.60    & 2.44 & 3.16    & 6.51   \\ \hline

\multicolumn{1}{l|}{\emph{Subject3}} & MEAN & -2.75  & 0.236   & -0.794 & -1.35  & 4.02   & 2.55    & 6.60 & 0.781   & -0.981 \\
\multicolumn{1}{l|}{}          & STD  & 2.41   & 1.20    & 3.21   & 3.84   & 4.76   & 3.66    & 1.93 & 0.936   & 2.34   \\ \hline
\multicolumn{1}{l|}{\emph{Subject4}} & MEAN & 3.11   & -1.83   & -2.87  & -5.53  & -2.07  & 2.65    & 4.54 & -0.491  & 0.714  \\
\multicolumn{1}{l|}{}          & STD  & 4.70   & 5.99    & 2.98   & 6.33   & 16.8   & 4.19    & 2.56 & 2.18    & 4.68   \\ \hline
\end{tabular}
}
 \caption{\textbf{Pose distribution of 4 sequences in the order of yaw, pitch, roll in degrees.} }
 \label{tbl:supp_dataset_analysis}
\end{table}

\section{Additional Results}
\label{sec:supp_results}

\begin{table}
\footnotesize

\begin{center}
\resizebox{1\linewidth}{!}{
\begin{tabular}{l|llll|llll|}
         & \multicolumn{4}{c|}{\emph{Subject1}} & \multicolumn{4}{c|}{\emph{Subject2}} \\

         & $\ell_1$ $\downarrow$ & SSIM  $\uparrow$ & LPIPS $\downarrow$ & PSNR $\uparrow$ & $\ell_1$ $\downarrow$   & SSIM $\uparrow$ & LPIPS$\downarrow$ & PSNR $\uparrow$ \\
\hline
Nerface   & 0.058          & 0.903          & 0.105          & 21.457          & 0.077          & {\color{blue}0.889}          & 0.121           & 18.252 \\
IM Avatar(-) & 0.068          & 0.901          & 0.113          & 20.502          & 0.093          & 0.877          & 0.157           & 14.960 \\
RigNeRF*   & {\color{blue}0.055}          & {\color{blue}0.904}          & 0.095          & {\color{blue}22.324}          & {\color{blue}0.072}          & 0.884          & 0.120           & {\color{blue}18.922} \\ 

RigNeRF*$\dag$   & 0.0767          & 0.876          & 0.108          & 18.829          & 0.0814          & 0.881         & 0.131          & 17.598 \\

\hline

Ours$\dag$      & 0.0773 & 0.895 & {\color{blue}0.0611} & 19.681 & 0.0873 & 0.876 & {\color{blue}0.0795} & 16.917 \\
Ours      & \textbf{0.054} & \textbf{0.929} & \textbf{0.051} & \textbf{23.508} & \textbf{0.062} & \textbf{0.917} & \textbf{0.0576} & \textbf{20.4375} \\

         & \multicolumn{4}{c|}{\emph{Subject3}} & \multicolumn{4}{c}{\emph{Subject4}} \\

         & $\ell_1$ $\downarrow$            & SSIM $\uparrow$          & LPIPS $\downarrow$         & PSNR $\uparrow$& $\ell_1$ $\downarrow$            & SSIM $\uparrow$          & LPIPS $\downarrow$         & PSNR $\uparrow$           \\
\hline
Nerface   & 0.047          & 0.894          & 0.055          & 22.220 & 0.077          & 0.818 & 0.085          & 17.910 \\
IM Avatar(-) &  0.043          & 0.900          & 0.078          & 23.218 & \textbf{0.063} & \textbf{0.870}          & 0.069          & {\color{blue}19.215} \\
RigNeRF*   & 0.035 & 0.910 & 0.052          & {\color{blue}24.634} & {\color{blue}0.065}          & {\color{blue}0.844}          & 0.063          & \textbf{19.253} \\
RigNeRF*$\dag$   & 0.03430 & 0.898 & 0.0526          & 23.109 & 0.118          & 0.727          & 0.129         & 14.677 \\
\hline

Ours$\dag$      & {\color{blue}0.0335}          & {\color{blue}0.925}          & {\color{blue}0.0387}  & 24.552 & 0.0745        & 0.789         & \textbf{0.0453} & 16.834    \\
Ours      & \textbf{0.0206}          & \textbf{0.971}          & \textbf{0.0265} & \textbf{30.854} & 0.081          & 0.830          & {\color{blue}0.062} & 19.085

\end{tabular}
}\vspace{-5pt}

    \end{center}
    \caption{Extended quantitative comparisons for~\cref{tab:comparisons_real}. Ours$\dag$ and RigNeRF*$\dag$ are run without optimized per-frame deformation latent code. \textbf{Bold black} is best result; {\color{blue} blue} is second best.}
    \label{tab:comparisons_real-extended}
\end{table}

\begin{figure*}[ht]
\begin{center}
 \includegraphics[width=0.9\linewidth]{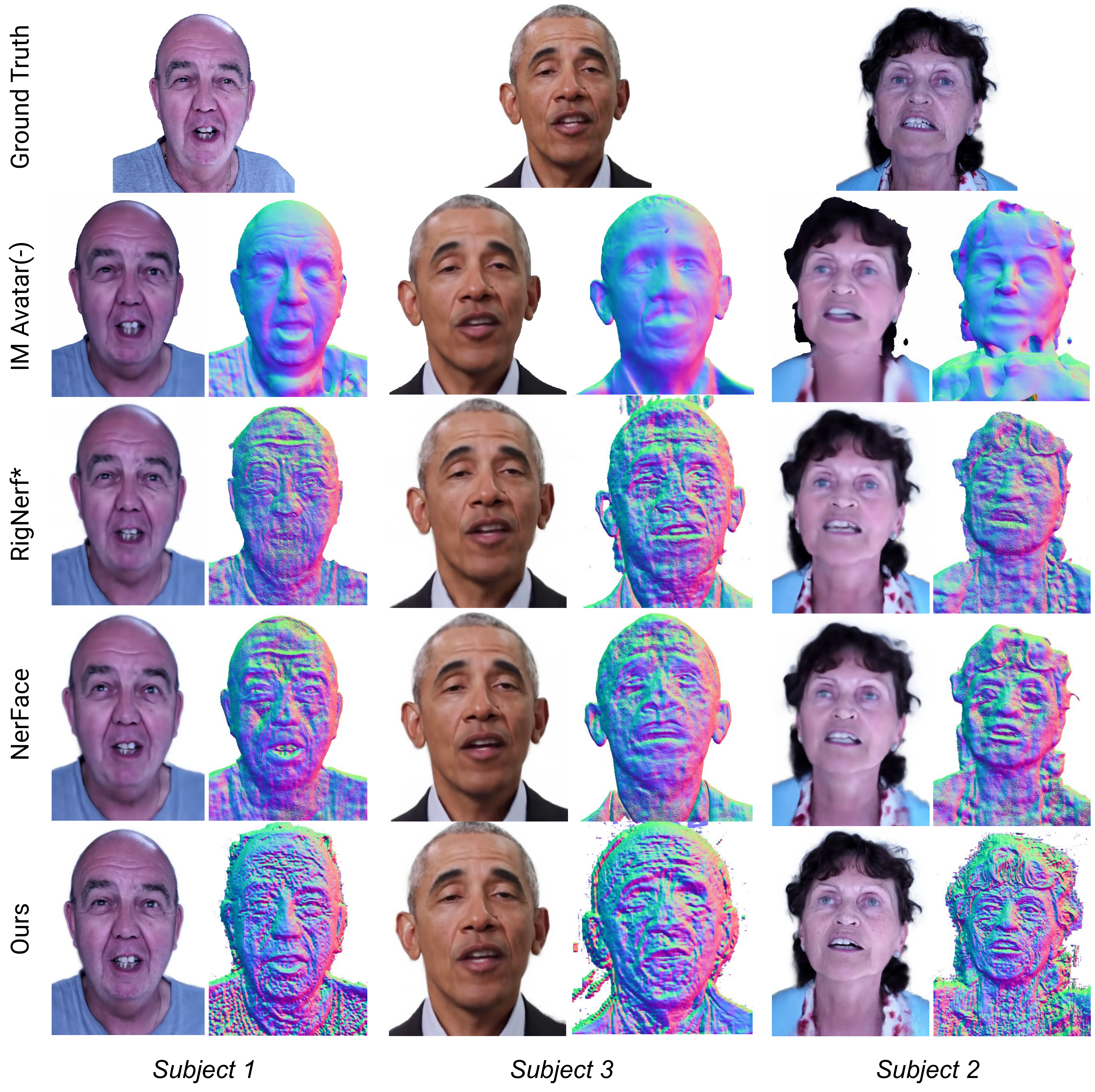}
\caption{\textbf{Qualitative comparisons of geometry with state-of-the-art on test data without tuning of latent code during test time.} Top to bottom: GT, IM Avatar(-), RigNeRF*, NerFace, and our approach. Here, the images are rendered with the latent codes of the first training frame of each sequence. Note that our approach produces significantly richer geometric details, as observed in the normal maps. Besides, the rendered images generated by our method faithfully reflect the pose, expression, and appearance of the ground truth images.}
\label{fig:comparisons_normal}
\end{center}
\end{figure*}


Test-time latent code optimization (deformation and appearance) helps adjust pose inaccuracies, which is a current limitation of our approach. However, our model excels at producing high-quality reconstruction, even without per-frame latent code optimization.
\cref{fig:comparisons_normal} compares the quality of the normals obtained by our methods and the different baselines on \emph{Subject1}, \emph{Subject2}, and \emph{Subject3}. Note that no image detail is lost when compared to~\cref{fig:comparison}. Besides, our method produces crisper results than baseline approaches, as shown in the reconstructed normal maps.

\cref{tab:comparisons_real-extended} shows additional metrics for Ours$\dag$ and RigNeRF*$\dag$, both using no optimized latent code. Ours$\dag$ suffers from global pose misalignment (several pixels off) without optimized latent code, impacting per-pixel image metrics (PSNR and $\ell_1$). Local facial structures are still well preserved as demonstrated by consistently lower scores (second best) on LPIPS for which we still achieve state-of-the-art performance.
\section{Extended Analysis}
\label{sec:supp_analysis}

\begin{figure*}[!ht]
\begin{center}
 \includegraphics[width=0.85\linewidth]{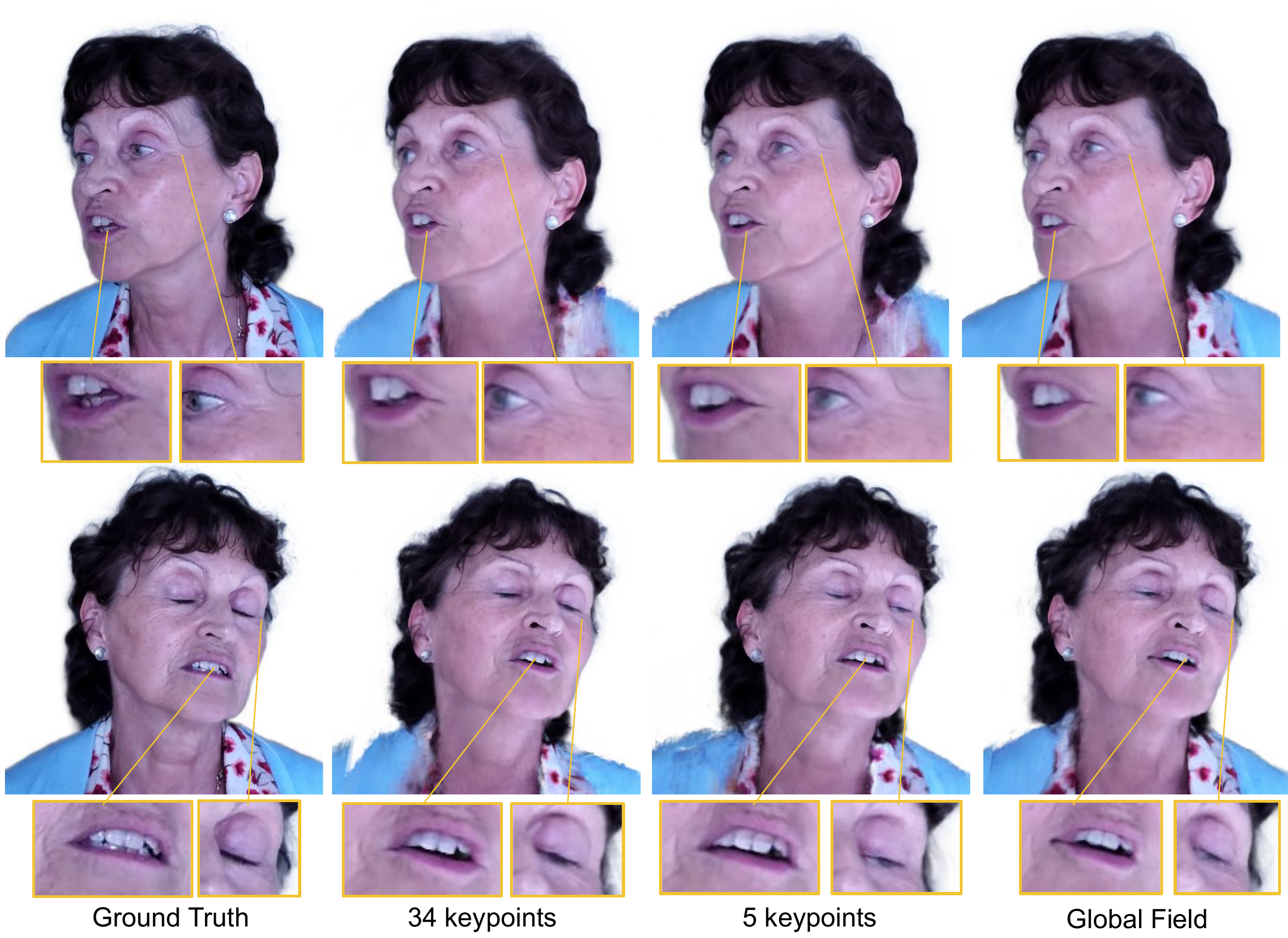}
\caption{\textbf{Ablation on the number of keypoints used by our method.} From left to right: ground truth, 34 keypoints, 5 keypoints, and global field (\ie, no local decomposition). Note that the network size for each experiment is adjusted such that the total number of parameters is roughly the same. Our method produces better visual results with 34 keypoints.}
\label{fig:comparisons_number_lmks}
\end{center}
\end{figure*}

\begin{table}[!hb]
    \begin{center}
    \begin{tabular}{ccc}
    \hspace{-12pt} \includegraphics[width=0.5\linewidth]{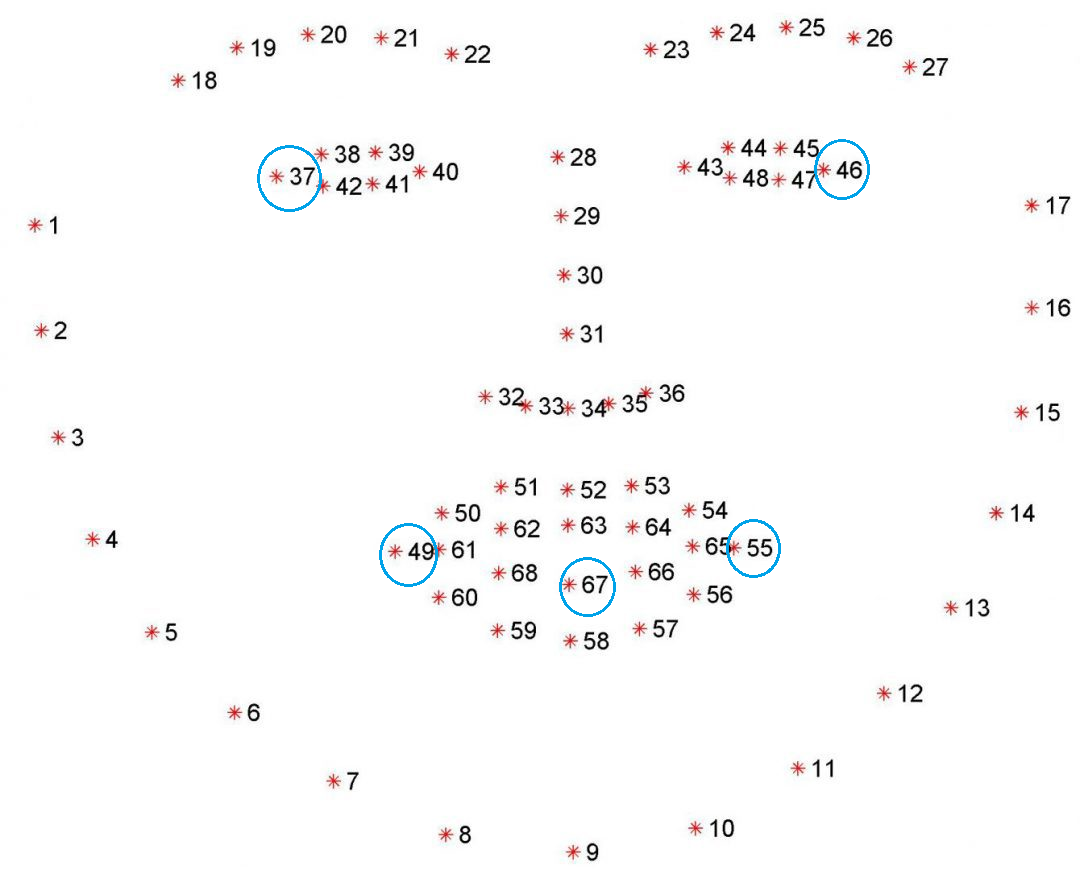} & \hspace{-12pt} \includegraphics[width=0.5\linewidth]{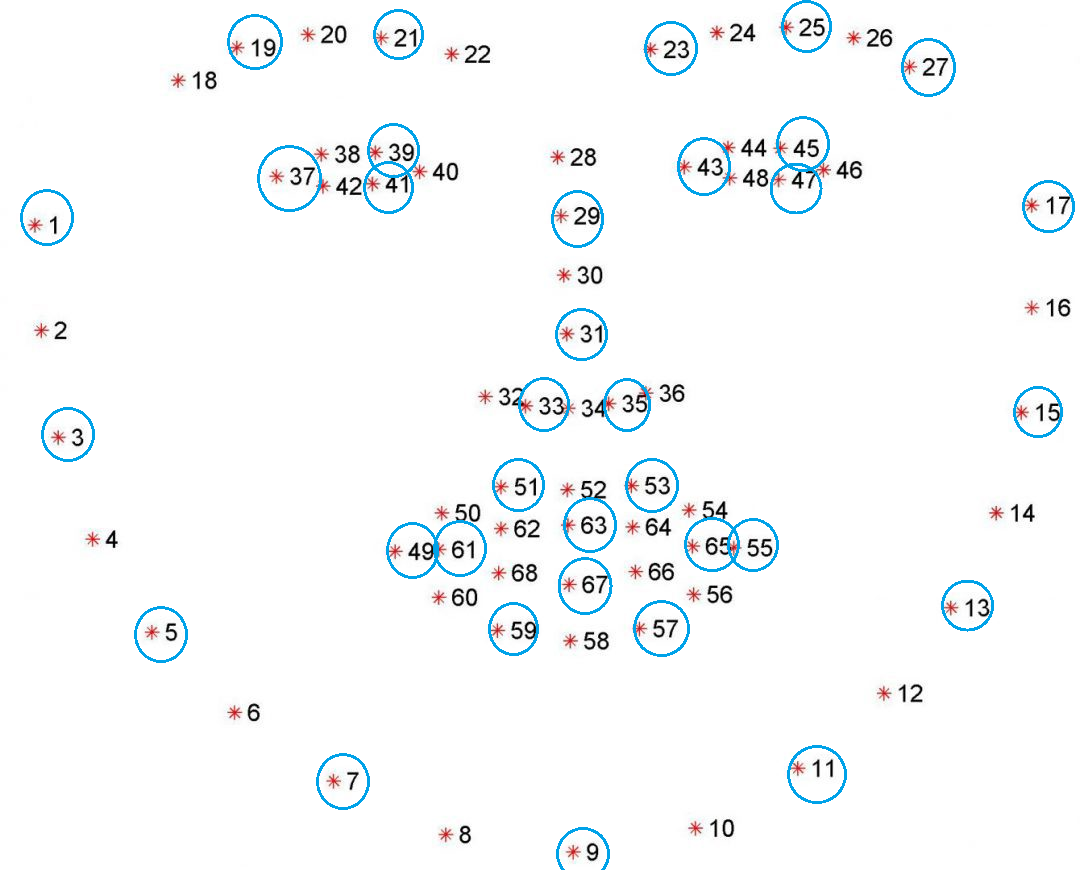}
    \end{tabular}
    \end{center}
    \caption{Keypoint locations for modeling local deformation fields. \emph{Left:} Locations of 5 keypoints. \emph{Right:} Locations of 34 keypoints.}
    \label{fig:lmk_vis}
\end{table}

\paragraph{Modeling Local Deformation Fields}
In our implementation, we model local deformation fields around a subset of facial landmarks using DECA's landmark definition \cite{feng2021_deca}. Specifically, we experimented with 5 and 34 keypoint locations, as shown in \cref{fig:lmk_vis}. The former representation is similar to \cite{zheng2022_imface}, though we change the tip of the nose with the midpoint of the lower lip to better model jaw deformations. In the latter, denser representation, we exploit the semantics of the facial landmark definition and center local fields around every other landmark.

\cref{fig:comparisons_number_lmks} compares the effect of using different numbers of keypoints. Note that our method reconstructs sharper details and more accurate facial features when the deformation field is decomposed with 34 keypoints. Overall local decomposition with both 34 and 5 keypoints results in better reconstruction of details than that of the global field. The latter tends to over smooth the reconstructed surface and produce less accurate facial deformations than multiple local deformation fields.

\end{document}



\newcommand{\mo}[1]{\textcolor{red}{[\texttt{Matt}: #1]}}
\newcommand{\pg}[1]{\textcolor{green}{[\texttt{Pablo}: #1]}}
\newcommand{\gb}[1]{\textcolor{blue}{[\texttt{Gaurav}: #1]}}
\newcommand{\sal}[1]{\textcolor{magenta}{[\texttt{Sally}: #1]}}

\newcommand{\todo}[1]{\textcolor{cyan}{[\texttt{TODO}: #1]}}

\newcommand{\isdraft}{false}


\title{Implicit Neural Head Synthesis via Controllable Local Deformation Fields\\
Supplementary Material}

\author{Chuhan Chen\\
Carnegie Mellon University\\
Institution1 address\\
{\tt\small chuhanc@andrew.cmu.edu}
\and
Matt O'Toole\\
Carnegie Mellon University\\
First line of institution2 address\\
{\tt\small mpotoole@cmu.edu}
\and
Gaurav Bharaj\\
Flawless AI\\
First line of institution2 address\\
{\tt\small gaurav.bharaj@flawlessai.com}
\and
Pablo Garrido\\
Flawless AI\\
First line of institution2 address\\
{\tt\small pablo.garrido@flawlessai.com}
}
\maketitle


\input{supplementary/01_method}

\input{supplementary/02_implementation}
\input{supplementary/03_datasets}
\input{supplementary/04_results}

{\small
\bibliographystyle{ieee_fullname}
\bibliography{main}
}